\documentclass[10pt,journal,compsoc]{IEEEtran}
\newif\ifpeerreview

\peerreviewfalse
\usepackage[nocompress]{cite}
\usepackage{url}
\usepackage{amsmath,amssymb,graphicx}
\usepackage{bm}
\usepackage{cleveref}
\DeclareMathOperator{\EX}{\mathbb{E}} %expected value
\usepackage{xcolor}
\usepackage{algpseudocode}
\usepackage{algorithm}

\usepackage{caption}
\usepackage{caption}

\usepackage{lipsum} % Only used to generate random text.
\crefname{table}{Table}{Tables} % This will make "Table" always appear with an uppercase "T".

\usepackage[absolute,overlay]{textpos}  % Enables absolute positioning

\usepackage[switch]{lineno}
\usepackage{enumitem}
\newcommand{\paperID}{17}

\title{NnD: Diffusion-based Generation of\\ 
Physically-Nonnegative Objects}

\author{Nadav Torem, Tamar Sde-Chen and Yoav Y. Schechner
 \IEEEcompsocitemizethanks{\IEEEcompsocthanksitem Nadav Torem, Tamar Sde-Chen and Yoav Y. Schechner are with the Viterbi Faculty of Electrical
and Computer Engineering, Technion-Israel Institute of Technology, Haifa
3200003, Israel (e-mail: torem@campus.technion.ac.il).}}

\begin{document}

\IEEEtitleabstractindextext{%
\begin{abstract}

Most natural objects have inherent complexity and variability. While some simple objects can be modeled from first principles, many real-world phenomena, such as cloud formation, require computationally expensive simulations that limit scalability. This work focuses on a class of physically meaningful, nonnegative objects that are computationally tractable but costly to simulate. To dramatically reduce computational costs, we propose nonnegative diffusion (NnD). This is a learned generative model using score based diffusion. It adapts annealed Langevin dynamics to enforce, by design, non-negativity throughout iterative scene generation and analysis (inference). NnD trains on high-quality physically simulated objects. Once trained, it can be used for generation and inference. We demonstrate generation of 3D volumetric clouds, comprising inherently nonnegative microphysical fields. Our generated clouds are consistent with cloud physics trends. They are effectively not distinguished as non-physical by expert perception. 
%This highlights the model’s ability to capture complex physical priors in a data-driven physically enforced framework.

\end{abstract}

}

% Make Title
\ifpeerreview
\linenumbers \linenumbersep 15pt\relax 
\author{Paper ID \paperID\IEEEcompsocitemizethanks{\IEEEcompsocthanksitem This paper is under review for ICCP 2025 and the PAMI special issue on computational photography. Do not distribute.}}
\markboth{Anonymous ICCP 2025 submission ID \paperID}%
{}
\fi
\maketitle

% The first section title should be wrapped inside a \IEEEraisesectionheading as follows.
\IEEEraisesectionheading{\section{Introduction}}
\label{sec:introduction}  
{

When objects are simple, they can be computationally modeled from first principles. For example, this is the case for made-man objects engineered using classic computed-aided design (CAD), or calculation of planetary motion. Moreover, in these cases, characterizing such objects based on measurements (an inverse problem) may sometimes be performed without statistical priors.
\begin{figure}[t]
    \centering
    \includegraphics[width=0.5\textwidth]{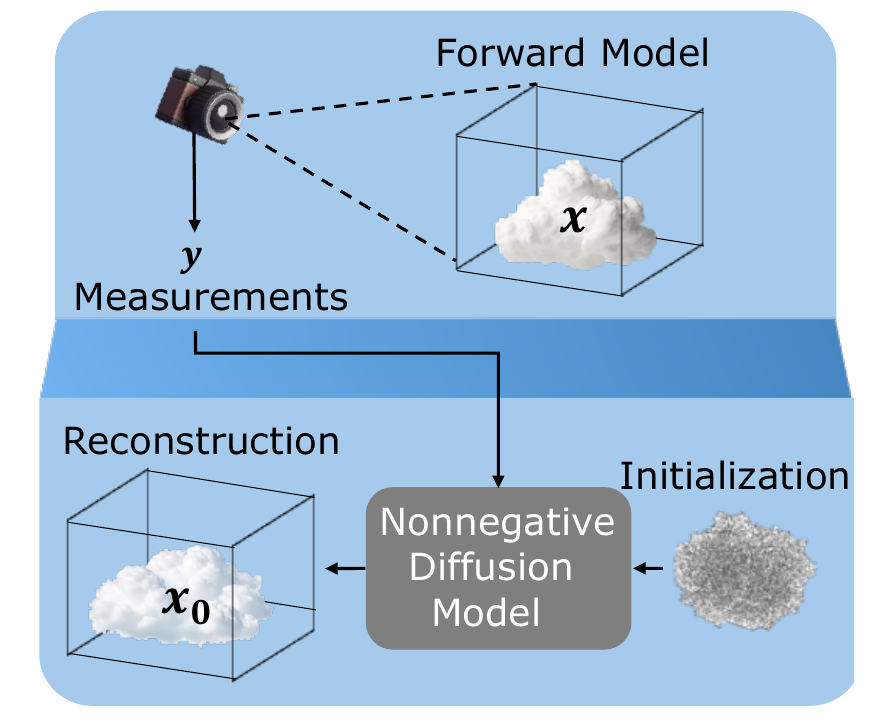}
    \caption{Starting from random noise, a learned diffusion model yields a generated object, or recovery based on measurements. The model is adapted to ensure, by design, nonnegative objects at each iteration. This makes both the object and the forward model compatible with nature. The forward model requires physically nonnegative objects.}
    \label{fig:Openin}
\end{figure}

On the other hand, most natural physical objects are complex, with significant variability, even when considering objects of a single class. Then, it is practically impossible to model and compute their creation from first principles. For example, there is no foreseeable digital computer that could mimic the entire array of microbiological and chemical processes that would eventually create a person in a simulation. This limitation applies to generation and inverse problems, e.g, seeking a physically-created object that corresponds to measurements. Hence, digital generation of such natural objects must be data-driven. The data are essentially samples from a high-dimensional probability distribution function (PDF), which expresses the prior on the object.
For example, digital generation of face images relies on sampling the prior. Furthermore, solving an imaging inverse problem for these objects (e.g., dehazing, deblurring, multiview~\cite{malik2023transient, mildenhall2021nerf, agrawal2023nova}) makes use of the data-driven prior. Actually, solving imaging inverse problems is often the only way to quantify these objects, for example in computed tomography.

There is an intermediate range of physical natural objects. While complex, they can still be modeled and computed from first principles of physical dynamics, at a high cost of resources (mainly time).  Digital simulations leading to creation of even a single object can require a super-computer and/or take weeks. The complexity of the simulations has a fundamental sequential nature, that cannot be mitigated by parallel computing. Thus, simulations are often of very small scales, take a long time to conclude, or generate relatively few results. 
Notable cases are objects whose physical generation is dictated by fluid dynamics. The basic (yet complicated) fields are derived by computational fluid dynamics as in water or air, including turbulence.
Complexity further increases 
where the flow has feedback with other object variables. These include: Material phase fields, as in clouds; 
Chemistry fields, as in air pollution dynamics and combustion; Electromagnetic fields, as in plasma;  Gravity fields relating to stars or nebulae.

The complexity motivates digital generation of such objects by {\em learned} models (Fig.~\ref{fig:Openin}), expressing their prior.  
Fortunately, computing such objects in high quality is feasible from first principles, even if expensively. Physical computations yield data to train a model to express a prior. 

This paper deals with {\em physically nonnegative} objects in the above complexity range. Often, sought objects are fundamentally nonnegative. Here are some examples.\\
%$\bullet~$Distance between object elements, e.g., interacting molecules in FRET, adjacent lenses in a barrel.\\
$\bullet~~$Size of object elements, e.g,, radius of scattering particles, thickness of coating layers.\\
%$\bullet~$Lifetime or half-life of object elements, e.g., molecular states in FLIM, nuclear spins in MRI.\\ 
$\bullet~$Temperature (above the absolute zero) of object elements, and resulting Doppler spectral broadening.\\
$\bullet~$Pressure (above vacuum) in an object element, and the speed of sound in the element.\\
$\bullet~$Density of particles in a volume element.\\
$\bullet~$Mass of an object element.\\
%$\bullet~$Energy above a stable equilibrium state (classic) in an object element. In quantum mechanical elements, e.g, vibrating molecules or electromagnetic fields, the energy level is strictly positive above the classic minimum.\\
%$\bullet~$Probability. Probability objects are created in light detection. There, radiance of an object element dictates a probability of photon emission, absorption or detection.\\
%$\bullet~$Some functions of probability, such as entropy.\\
Additional examples are listed in Sec.~\ref{sec:discuss}.  Nonnegativity has several aspects. First, physically, the objects are only defined and meaningful as such. Second, in some cases, there is no meaningful {\em forward model} that can operate on a negative object: there is no mathematical model relating a negative object to measurements. For example, there is no model for Doppler broadening that can relate spectral data to a negative temperature. 

A highly successful approach to a data-driven model for expressing a prior is a diffusion model. Diffusion offers stochastic generation with mathematical guarantees. Diffusion is utilized in various ways, including image generation~\cite{song2019generative,ho2020denoising,yuan2023physdiff,mikuni2023fast}, inverse problems~\cite{torem2023complex,chung2022diffusion, song2023pseudoinverse,feng2023score}, text-to-image conversion~\cite{rombach2022high,saharia2022photorealistic} and video synthesis~\cite{esser2023structure,blattmann2023align, zhang2023i2vgen}. We use a score-based diffusion approach titled {\em annealed Langevin dynamics} (ALD). ALD is not guaranteed to yield nonnegative results while it iterates. Thus, in this paper, we adapt ALD to nonnegative objects. We formulate nonnegative generation and inference that samples a highly probable object from the posterior PDF, given measurements. 

We demonstrate generation of 3D volumetric clouds. They are complex objects that are physically created in turbulent flow. The water droplet microphysical parameters per voxel must be nonnegative. A forward model relating them to images relies on non-negative droplet sizes. The results seem sufficient to 
%confuse 
convince a leading expert in cloud physics, that diffusion-based clouds are not distinguished as fake, relative to physics-based clouds. 

}

%%%%%%%%%%%%%%%%%%%%%%%%%%%
\section{Theoretical Background}
\label{sec:back}

%%%%%%%%%%%%%%%%%%5
\subsection{Langevin Dynamics}
\label{sec:LD}

Consider a true physical object $\bm{x}$. It is randomly sampled from nature, with a natural PDF denoted $p(\bm{x})$. We do not have hold of this object. We need to digitally generate an object (ideally, a natural one), by a process. A generated object is denoted $\bm{x}_0$.
Generation seeks to maximize the probability $p(\bm{x}_0)$. 

In the {\em Langevin dynamics} algorithm, maximization of $p(\bm{x}_0)$ is sought by a finite number of iterations, $T$. Let $t\in[T,\dots,1]$ be an iteration count-down index. Hence, a generated object at iteration $t$ is $\bm{x}_t$, and the process is initialized by an arbitrary unnatural object $\bm{x}_T$.
Let $\bm{I}$ be the identity matrix and $\bm{\eta}_t \sim \mathcal{N}(0, \bm{I})$ be white Gaussian noise at iteration $t$. Then, generation by Langevin dynamics 
follows
\begin{equation}
    \bm{x}_{t-1} = \bm{x}_t + \alpha_t\nabla_{\bm{x}_t}\log p(\bm{x}_{t}) + \sqrt{2\alpha_t}\bm{\eta}_t\;.
\label{eq:LD_gen1}
\end{equation}
Here $\alpha_t$ is a %an appropriately chosen 
small non-negative value that decays with $t$, such that $\alpha_{t\rightarrow 1}= 0$.
Intuitively, Eq.~(\ref{eq:LD_gen1}) performs gradient ascent of the log-probability, gradually increasing the probability of $\bm{x}_t$. The added random $\bm{\eta}_t$ helps avoiding degeneration to a local maximum. 
%stochastic sampling, avoiding a collapse to a local maximum of the distribution. 

A generalization of this process solves inverse problems, as common in imaging. Let object $\bm{x}$ be imaged. The corresponding measured image data is noisy and denoted $\bm{y}$. Consider the posterior PDF of $\bm{x}$, given $\bm{y}$, which is denoted
$p({\bm{x}}|\bm{y})$. Now, solving an inverse problem is a generative process, which  seeks to maximize $p({\bm{x}}|\bm{y})$. In analogy to Eq.~(\ref{eq:LD_gen1}), the process is 
\begin{equation}
    {\bm{x}}_{t-1} = {\bm{x}}_t + \alpha_t\nabla_{{\bm{x}}_t}
    \log p({\bm{x}}_{t}|\bm{y}) + \sqrt{2\alpha_t}\bm{\eta}_t\;.
\label{eq:LD_inverse}
\end{equation}
Following Bayes rule,
\begin{equation}
  \nabla_{{\bm{x}}_t}\log{p({\bm{x}}_t|\bm{y})} = \nabla_{{\bm{x}}_t}\log{p({\bm{x}}_t)} 
 + \nabla_{{\bm{x}}_t}\log{p(\bm{y}|{\bm{x}}_t)}  \;.
\label{eq:bayes}
\end{equation}
From Eqs.~(\ref{eq:LD_inverse},\ref{eq:bayes}),
\begin{equation}
    {\bm{x}}_{t-1} = {\bm{x}}_t + 
     \alpha_t\nabla_{{\bm{x}}_t}\log{p({\bm{x}}_t)} 
    +\alpha_t\nabla_{{\bm{x}}_t}\log{p(\bm{y}|{\bm{x}}_t)} 
    + \sqrt{2\alpha_t}\bm{\eta}_t.
\label{eq:LDML}
\end{equation}

The PDF $p(\bm{y}|{\bm{x}}_t)$ is the data likelihood. It is often well modeled. In the absence of noise, data $\bm{y}$ are  set by a {\em forward model},
$\bm{y}= \mathcal{F}\{\bm{x}\}$. The likelihood PDF is then 
spread by the imaging noise model. The noise model is often  known (e.g, Poissonian photon noise). Thus, a major unknown in both 
Eqs.~(\ref{eq:LD_gen1},\ref{eq:LDML}) is the function
$\nabla_{\bm{x}_t}\log p(\bm{x}_t)$.

%These measurements are derived from $\bm{x}$ through the \emph{Forward Model} $\mathcal{F}(\bm{x})$. Equivalently to Eq.~(\ref{eq:LD_gen}), the posterior distribution $p(\bm{x}|\bm{y})$ can be sampled from using the following transition rule,

%%%%%%%%%%%%%%%%%%%%%%%%%%%%
\subsection{Approximating a Score Function}
\label{sec:scoregen}

The \textit{score function} of a PDF is $\nabla_{\bm{x}}\log p(\bm{x})$. 
Equations~(\ref{eq:LD_gen1},\ref{eq:LDML}) require computation of the score function at each iteration. Unfortunately, we do not have access to $\nabla_{\bm{x}}\log p(\bm{x})$. To address this limitation, Ref.~\cite{song2020sliced} introduces an estimator, known as the \textit{score network}. 

The score network is a deep neural network (DNN), having parameters $\bm{\phi}$. The input to the score network is an object $\bm{x}$. The network outputs a function  
$s_{\bm{\phi}}(\bm{x})$. The DNN is trained so that 
$s_{\bm{\phi}}(\bm{x})$ approximates the score function $\nabla_{\bm{x}}\log p(\bm{x})$. Training is supervised, using true objects $\bm{x}$, which are natural samples of $p(\bm{x})$. Training minimizes a loss function $\cal C$.
A desired loss is of the form 
$\|s_{\bm{\phi}}(\bm{x})- \nabla_{\bm{x}}\log p(\bm{x})\|_2^2$. In practice, Ref.~\cite{song2020sliced} proposes a different loss, whose minimization is equivalent, i.e., yielding the same $\bm{\phi}$.

% Moreover, the loss function used to train the network is generally a tractable analytical term, which, up to a constant, is equivalent to minimizing the loss from the score function~\cite{song2019generative}.

Estimation using a score network has limitations. First, the score network $s_{\bm{\phi}}(\bm{x})$ is not scalable to high dimensional objects, due to a costly computation~\cite{song2019generative,song2020sliced} of the loss function $\cal C$.
Second, $s_{\bm{\phi}}(\bm{x})$ is unreliable for objects having a low  probability $p(\bm{x})$. There are two reasons for this. (i) Hardly any training samples exist of objects having a low  $p(\bm{x})$, thus under-constraining the DNN. (ii) In typical PDFs, the gradient $\nabla_{\bm{x}}\log p(\bm{x})$ is very low, at and around $\bm{x}$ for which $p(\bm{x})$ is low. Consequently, Eqs.~(\ref{eq:LD_gen1},\ref{eq:LDML}) do not evolve properly. To address these limitations, Ref.~\cite{song2019generative} introduces \textit{annealed Langevin dynamics} (ALD), described next.

%%%%%%%%%%%%%%%%%%%%%
\subsection{Annealed Langevin Dynamics}
\label{sec:ALD}

To overcome low $p(\bm{x})$ values, the PDF can be blurred as a function of the object $\bm{x}$. Blurring a PDF bleeds high probability values from highly probable objects $\bm{x}$ to objects ${\bm{x}}_t$ encountered in the generation process, which generally have a low $p{({\bm{x}}_t)}$. The blurred PDF is denoted $p_{\tilde{\bm{x}}}(\tilde{\bm{x}})$.

A way to blur a PDF is to add a random vector (noise) $\bm{n}$ to a true object $\bm{x}$. Then, the true-object PDF is convolved with the PDF of the noise. Hence define
an {\em annealed} object 
\begin{equation}
    \tilde{\bm{x}}_t =  {\bm{x}} + \bm{n}_t
    \;
\label{eq:blurred}
\end{equation}
at iteration $t$, where $\bm{n}_t\sim\mathcal{N}(0, \sigma^2_t\bm{I})$. Generative iterations then evolve 
$\tilde{\bm{x}}_t$. Initially, being far from a natural object, the variance $\sigma^2_T$ is high, meaning that 
$\tilde{\bm{x}}_T$ is a very noisy scene. The blurred PDF 
$p_{\tilde{\bm{x}}}(\tilde{\bm{x}}_T)$ is nevertheless effective  at $\tilde{\bm{x}}_T$, leading to iterative refining of the scene. This refinement is akin to reversing a noise process. Hence, as iteration indices count-down from $T$ to 1, $\sigma^2_t$ decays. %Consequently, at the final iterations, the score function  $\nabla_{\bm{x}}\log {\tilde p}(\tilde{\bm{x}}_t)$ resembles  $\nabla_{\bm{x}}\log p(\tilde{\bm{x}}_t)$. 
This process is ALD. Rather than Eqs.~(\ref{eq:LD_gen1},\ref{eq:LDML}), it uses
\begin{equation}
    \tilde{\bm{x}}_{t-1} = \tilde{\bm{x}}_t + \alpha_t\nabla_{\tilde{\bm{x}}_t}\log p_{\tilde{\bm{x}}}(\tilde{\bm{x}}_t) + \sqrt{2\alpha_t}\bm{\eta}_t\;,
\label{eq;ALD_gen}
\end{equation}
\begin{align}
    \tilde{\bm{x}}_{t-1} = \tilde{\bm{x}}_t + 
     \alpha_t\nabla_{\tilde{\bm{x}}_t}
     \log{p_{\tilde{\bm{x}}}(\tilde{\bm{x}}_t)} 
    +\alpha_t\nabla_{\tilde{\bm{x}}_t}\log{p(\bm{y}|\tilde{\bm{x}}_t)}\nonumber \\
    + \sqrt{2\alpha_t}\bm{\eta}_t.
\label{eq:ALDML}
\end{align}
The parameter $\alpha_t$ should then be set per $t$ in accordance with $\sigma_t$. The true $\sigma_t$ is unknown, since the true ${\bm{x}}$ in Eq.~(\ref{eq:blurred}) is unknown.
Thus, an assumed step-wise function $\sigma_t$ is assumed: it is part of the hyper-parameter set of the algorithm. 

In analogy to Sec.~\ref{sec:scoregen}, the score function
$\nabla_{\bm{x}}\log{p_{\tilde{\bm{x}}}(\tilde{\bm{x}}_t)}$ 
is approximated by the output 
$s_{\bm{\theta}}(\tilde{\bm{x}}_t)$ of a trained DNN. The DNN parameters are ${\bm{\theta}}$. However, the training cost $\tilde{\cal C}$ is analytically tractable, being computationally more affordable for high-dimensional objects. 

We stress that annealing noise in Eq.~(\ref{eq:blurred}) is conceptual. It is not actual noise that is intentionally added during most iterations of the generative process. The only implications of annealing are; initialization by a very noisy scene $\tilde{\bm{x}}_T$; the function $\alpha_t$, which based on an assumed $\sigma_t$, and a DNN  $s_{\bm{\theta}}(\tilde{\bm{x}}_t)$ approximating a score function of a blurred PDF.

\subsection{Imaging}
\label{sec:forward_model}

In the absence of noise, as written in Sec.~\ref{sec:LD}, $\bm{y}=\mathcal{F}\{\bm{x}\}$. 
Denote incorporation of noise into the expected signal by the operator
$\mathcal{W}$ . Then, raw measured data satisfies
\begin{equation}
    \bm{y} = \mathcal{W}[\mathcal{F}\{\bm{x}\}]
    \;.
    \label{eq:datanoise}
\end{equation}
We have so far referred to a generic relation between $\bm{x}$ and $\bm{y}$. Now, we focus on imaging. In optical imaging, radiometric measurements are typically dominated by photon noise, which is Poissonian and independent per datum (pixel, time sample etc.). Let the radiometric units of the data and the forward model be in photo-electrons. Then, in photon noise, the variance equals $\mathcal{F}\{\bm{x}\}$. To ease computations, photon noise per datum is often approximated as Gaussian, with variance $\mathcal{F}\{\bm{x}\}$, i.e.,
\begin{equation}
    \bm{y} \approx \mathcal{F}\{\bm{x}\} +
    \mathcal{N}(0, \mathcal{F}\{\bm{x}\})
    \;.
    \label{eq:normalnoise}
\end{equation}

In inverse problems, $\mathcal{F}\{\bm{x}\}$ is unknown, because $\bm{x}$ is unknown. Therefore,
the noise in Eq.~(\ref{eq:normalnoise}) is approximated as 
$\mathcal{N}(0, \bm{y})$. Hence, the likelihood for the $k$'th datum element (pixel, viewpoint, or wavelength sample) $y_k$ is approximately 
\begin{equation}
\label{eq:p_y_xk}
\begin{aligned}
      p(y_k|\bm{x}) \approx 
      \frac{1}{\sqrt{2\pi y_k}}
      \exp\left[{-\frac
      {(y_k - \mathcal{F}_k\{\bm{x} \})^2}{2 y_k}}\right],
\end{aligned}
\end{equation}
where $\mathcal{F}_k\{\bm{x}\}$ is the corresponding  noiseless model value. 
% \begin{equation}
% \label{eq:p_y_x}
% \begin{aligned}
%       p(\bm{y}|\bm{x}) \approx 
%       \frac{1}{\sqrt{2\pi \bm{y}}}
%       \exp\left[{-\frac
%       {(\bm{y} - \mathcal{F}\{\bm{x} \})^2}{2\bm{y}}}\right],
% \end{aligned}
% \end{equation}

Typically in imaging, the forward model is a differential rendering operator. Image rendering expresses light propagation, reflection, refraction, scattering, absorption, motion and defocus blur, spatiotemporal and spectral integration and sampling in a camera.  Differential rendering is often well known and numerically implemented. However, rendering can become complex. Complexity of rendering is caused by multiple (high order) scattering and/or multiple reflections. These lead to a recursive nonlinear relation of the data to the object. Hence, in these cases, there is high computational cost on operations of the forward model and on the diffusion (differential) steps. 

\section{The Problem}
\label{sec:prob}

The process of Sec.~\ref{sec:ALD} is unsuitable for generating objects which are strictly nonnegative. Here are the reasons.\\
$\bullet$~ The process initialization $\tilde{\bm{x}}_T$ in 
Sec.~\ref{sec:ALD} is strong Gaussian noise having zero mean. Consequently, about half of the elements in 
$\tilde{\bm{x}}_T$ violate nonnegativity. \\
$\bullet$~ In Eqs.~(\ref{eq;ALD_gen},\ref{eq:ALDML}), the intentionally injected noise $\bm{\eta}_t$ is Gaussian. Hence, generally, this noise  shifts some elements of  
$\tilde{\bm{x}}_{t-1}$ to a forbidden negative domain.\\
$\bullet$~ Equation~(\ref{eq:blurred}) includes Gaussian noise. So, if the annealed scene $\tilde{\bm{x}}_t$ is nonnegative, then the true scene ${\bm{x}}$ generally has negative elements, which is impossible.\\
$\bullet$~ On the other hand, since by definition ${\bm{x}}$ is nonnegative, then the annealed scene $\tilde{\bm{x}}_t$ generally has negative elements. However, this situation in incompatible with the likelihood function $p(\bm{y}|\tilde{\bm{x}}_t)$ appearing in 
Eq.~(\ref{eq:ALDML}). The reason is that likelihood computation relies on a forward model  
$\mathcal{F}(\tilde{\bm{x}}_t)$, as described in Sec.~\ref{sec:LD}. Often, the forward model is undefined for a negative object. For example, there is no phase function for Mie scattering from a particle having a negative radius. No image formation model is defined for negative particle radii. A forward model expects valid physical inputs; not necessarily probable ones, allowing for noisy scenes that may be improbable, but nevertheless possible.

\section{NnD: Nonnegative Diffusion of Objects}
\label{sec:nonngeg}

% At each iteration $t$, a noisy scene ${\tilde{\bm{{x}}}_t}$ is generated. In a diffusion process, the Markov chain does not constrain the values at each point in space (e.g., pixels or voxels). Instead, the diffusion process focuses on the distribution of ${\tilde{\bm{{x}}}_t}$, as defined in Eq.~(\ref{eq:blurred}). The blurred annealing approach, inherently results in non-physical annealing scenes  ${\tilde{\bm{{x}}}_t}$.

% For example, in the naive score-based approach, initializing ${\tilde{\bm{{x}}}_T} \sim \mathcal{N}(0, \sigma^2_0)$ statistically results in half of the scene containing negative values (e.g., negative photon-electron counts).
% Some methods introduce conditioned or constrained diffusion processes. While these approaches may eventually produce a solution %${\bm{{x}}_0}$%
% that satisfies the necessary constraints, the core issue remains: at each diffusion step, a noisy scene ${\tilde{\bm{{x}}}_t}$ passes through a physical forward model. The forward model expects valid physical inputs (not necessarily probable ones, allowing for noisy scenes that are improbable). Hence, the annealing process defined in Eq.~(\ref{eq:blurred}) can not be used in physical constrained problems, where the forward model expects physically valid values.
%(ii) When $lwc$ equals zero, $r_{eff}$, $v_{eff}$ must also equal zero. 

We now address the problem detailed in Sec.~\ref{sec:prob}. A possible solution is to add nonnegative (non Gaussian) noise, or clip  values. So far, we found this to yield unsatisfactory results. One reason, we believe, is that DNNs that are surrogate to a score function tend to learn better on Gaussian noise. 
%While alternative positive noise distributions, such as Log-Normal, Poisson, or shifted Folded Gaussian distributions, could be considered, notable challenges arise. Firstly, analytically computing the gradient term $\nabla_{\tilde{\bm{{x}}}_t}\log{p(\tilde{\bm{{x}}}_t)}$ may not be achievable without imposing strong assumptions. Secondly, deep neural networks often struggle to learn $\nabla_{\tilde{\bm{{x}}}_t}\log{p(\tilde{\bm{{x}}}_t)}$ for more complex noise distribution due to there inherent complexity and nonlinear behavior. 
Therefore, we choose to work within the well established Gaussian distribution framework. 
%, which is optimal for a minimum mean squared error (MMSE) denoiser. 
%In our upcoming work, we plan on applying the diffusion process in an analytically tractable latent space, allowing us to use both the Gaussian framework and ease mathematical derivations.   
To do so, we define a latent field 
$\bm{\rho}$, related by a point-wise exponent to 
$\bm{x}$,
\begin{equation}
\label{eq:x_rho}
    \bm{x} = \exp(\bm{\rho}) \;.
\end{equation}
 Then, diffusion progresses in the latent space, $\bm{\rho}$, which can have negative elements. This yields a generated field $\bm{\rho}_0$. Then, the generated nonnegative scene is 
\begin{equation}
    \bm{x}_0 = \exp(\bm{\rho}_0) \;.
    \label{eq:x0rho0}
\end{equation}

Based on a natural nonnegative object $\bm{{x}}$, define an annealed real-valued field $\tilde{\bm{\rho}}_t $, using a point-wise logarithm,
\begin{equation}
    \tilde{\bm{\rho}}_t = \log{({\bm{{x}}} +\epsilon)} + \mathcal{N}(0, \sigma^2_t)
    \;.
    \label{eq:tilderhot}
\end{equation}
Here $\epsilon$ %=  \exp\left(-10\right)$ to the log term to 
is a small value, introduced to numerically stabilize the calculations for scene elements having null values.
% The definition of $\rho$ helps us in two ways: (i) It holds positive values only for cloud generation. (ii) the term $\exp{\rho}$ is
Then, preceding the step of Eq.~(\ref{eq:x0rho0}), ALD for scene generation or recovery take, respectively, the forms
\begin{equation}
    \tilde{\bm{\rho}}_{t-1} =  \tilde{\bm{\rho}}_{t} + \alpha_t\nabla_{\tilde{\bm{\rho}}_t}\log p(\tilde{\bm{\rho}_t}) + \sqrt{2\alpha_t}\bm{\eta}_t\;.
\label{eq:LD_gen_phy}
\end{equation}
\begin{align}
    \tilde{\bm{\rho}}_{t-1} = \tilde{\bm{\rho}}_t + 
     \alpha_t\nabla_{\tilde{\bm{\rho}}_t}
     \log{p_{\tilde{\bm{\rho}}}(\tilde{\bm{\rho}}_t)} 
    +\alpha_t\nabla_{\tilde{\bm{\rho}}_t}\log{p(\bm{y}|\tilde{\bm{\rho}}_t)}\nonumber \\
    + \sqrt{2\alpha_t}\bm{\eta}_t.
\label{eq:ALDrhoML}
\end{align}
We now describe how to compute the gradient of the prior term
$\nabla_{\tilde{\bm{\rho}}_t}\log{p_{\tilde{\bm{\rho}}}(\tilde{\bm{\rho}}_t)}$ and the likelihood gradient term 
$\nabla_{\tilde{\bm{\rho}}_t}\log{p(\bm{y}|\tilde{\bm{\rho}}_t)}$.

\subsection{Gradient of the Prior Term}
\label{sec:gradprior}

First, we describe how $\nabla_{\tilde{\bm{\rho}}_t}\log{p_{\tilde{\bm{\rho}}}(\tilde{\bm{\rho}}_t)}$ is estimated. We are motivated by prior work on diffusion models of real-valued objects, and make appropriate adaptations for the latent space.
For notation brevity, from now on in this paper, the operations $(\cdot)^2,\sqrt{\cdot},\exp[{\cdot}]$, multiplication and division use vector form, but mean point-wise operations per vector element. 
Based on Eq.~(\ref{eq:tilderhot}), 
%the posterior distribution, $p(\tilde{\bm{\rho}}_t|\bm{x})$, can be analytically traced. Namely, 
\begin{equation}
            p(\tilde{\bm{\rho}}_t|\bm{{x}})
         = 
         %{p \left(\mathcal{N}(0, \sigma^2_t)|\bm{{x}}\right)}=
        \frac{1}{\sqrt{2\pi\sigma^2_t}}\exp\left[{\!-\frac{(\tilde{\bm{\rho}}_t - {\log{(\bm{x} + \epsilon)}})^2}{2\sigma^2_t }}\right]
         \;.
        \label{eq:px_tiled_O}
\end{equation}

By definition, the marginal distribution  $p(\tilde{\bm{\rho}}_t)$ satisfies
\begin{equation}
    p(\tilde{\bm{\rho}}_t) = \int p(\tilde{\bm{\rho}}_t|\bm{x})p(\bm{x})d{\bm{x}}
     \;.
    \label{eq:po_tiled}
\end{equation}
Inserting Eq.~(\ref{eq:px_tiled_O}) into Eq.~(\ref{eq:po_tiled}) yields,
\begin{equation}
    p(\tilde{\bm{\rho}}_t) = \int  \frac{1}{\sqrt{2\pi\\\sigma^2_t}}\exp\left[{\!-\frac{(\tilde{\bm{\rho}}_t - {\log{(\bm{x} + \epsilon)}})^2}{2\sigma^2_t }}\right]p(\bm{x})d{\bm{x}}
     \;.
    \label{eq:po_tiled2}
\end{equation}
The gradient  of Eq.~(\ref{eq:po_tiled2})  satisfies
\begin{gather}
        \nabla_{\tilde{\bm{\rho}}_t}p(\tilde{\bm{\rho}}_t) %= \int \frac{1}{\sqrt{2\pi\\\sigma{^2}_t}}\nabla_{\tilde{\bm{\rho}}_t}\exp\left[{\!-\frac{(\tilde{\bm{\rho}}_t - {\log{(\bm{x} + \epsilon)}})^2}{2\sigma^2_t }}\right]p(\bm{x})d{\bm{x}} 
        %\nonumber\\
        = \int \left[\frac{ {\log{(\bm{x} + \epsilon)}}-\tilde{\bm{\rho}}_t}{\sigma^2_t}\right]
        p(\tilde{\bm{\rho}}_t|\bm{x})p(\bm{x})d{\bm{x}}
     \;.
     \label{eq:grad_po1}
\end{gather}
% Inserting the definition of $p(\tilde{\bm{\rho}}_t|\bm{x})$ given in \cref{eq:px_tiled_O} herein to \cref{eq:grad_po1} herein results in
% \begin{equation}
%     \nabla_{\tilde{\bm{\rho}}_t}p(\tilde{\bm{\rho}}_t) = \int\left[\frac{ \bm{x}-\tilde{\bm{\rho}}_t}{\sigma^2_t}\right]p(\tilde{\bm{\rho}}_t|\bm{x})p(\bm{x})d\bm{x}\;.
%     \label{eq:Ograd_result}
% \end{equation}

Divide both sides of Eq.~(\ref{eq:grad_po1})  by $p(\tilde{\bm{\rho}}_t)$. This leads to
\begin{equation}
\frac{\nabla_{\tilde{\bm{\rho}}_t}p(\tilde{\bm{\rho}}_t)}{p(\tilde{\bm{\rho}}_t)} = \int\left[\frac{ {\log{(\bm{x} + \epsilon)}}-\tilde{\bm{\rho}}_t}{\sigma^2_t}\right]\frac{p(\tilde{\bm{\rho}}_t|\bm{x}){p(\bm{{x}})}}{p(\tilde{\bm{\rho}}_t)}d\bm{x}
\;.
\label{eq:Oalmost}
\end{equation}
Recall Bayes rule, 
\begin{equation}
    p(\bm{x}|\tilde{\bm{\rho}}_t) = \frac{p(\tilde{\bm{\rho}}_t|\bm{x})p(\bm{x})}{p(\tilde{\bm{\rho}}_t)}\;.
    \label{eq:bayes3}
\end{equation}
Inserting Eq.~(\ref{eq:bayes3}) into Eq.~(\ref{eq:Oalmost}) results in
\begin{equation}
     \frac{\nabla_{\tilde{\bm{\rho}}_t}p(\tilde{\bm{\rho}}_t)}{p(\tilde{\bm{\rho}}_t)}= \int\left[\frac{{\log{(\bm{x} + \epsilon)}}-\tilde{\bm{\rho}}_t}{\sigma^2_t}\right]p(\bm{x}|\tilde{\bm{\rho}}_t)d\bm{x}
     \;.
     \label{eq:Ofinal1}
\end{equation}
The left-hand side of Eq.~(\ref{eq:Ofinal1}) satisfies
\begin{equation}
    \frac{\nabla_{\tilde{\bm{\rho}}_t}p(\tilde{\bm{\rho}}_t)}{p(\tilde{\bm{\rho}}_t)} = \nabla_{\tilde{\bm{\rho}}_t}\log p(\tilde{\bm{\rho}}_t)\;.
    \label{eq:Ofinal2}
\end{equation}
%The right-hand side  of 
From Eqs.~(\ref{eq:Ofinal1},\ref{eq:Ofinal2}),  
%herein can be given by, 
\begin{equation}
    \begin{aligned}
        \nabla_{\tilde{\bm{\rho}}_t}\log p(\tilde{\bm{\rho}}_t)=
        \int\left[\frac{{\log{(\bm{x} + \epsilon)}}-\tilde{\bm{\rho}}_t}{\sigma^2_t}\right]&p(\bm{x}|\tilde{\bm{\rho}}_t)d\bm{x} 
        % \\&\EX\left[\frac{ {\log{(\bm{x} + \epsilon)}}-\tilde{\bm{\rho}}_t}{\sigma^2_t}\bigg|\tilde{\bm{\rho}}_t\right]
        \;.
    \label{eq:Ofinal3}
    \end{aligned}
\end{equation}
% Using the definition of \cref{eq:defining_x_tilde}, we remain with, 
% \begin{equation}
%     \EX\left[\frac{ \bm{x}-\tilde{\bm{\rho}}_t}{\sigma^2_t_t\bm{x}}|\tilde{\bm{\rho}}_t\right] = \EX\left[\frac{\bm{e}^{\rm annea}_t}{\sigma^2_t_t\bm{\sqrt{x}}}\right]
% \label{eq:final4}
% \end{equation}

Both $\tilde{\bm{\rho}}_t$ %is given, 
and $\sigma_t$ %, \epsilon$ 
are independent of $\bm{x}$, thus taken out of the integral. By definition,
$\int p(\bm{x}|\tilde{\bm{\rho}}_t)d\bm{x}=1$.
Consequently, Eq.~(\ref{eq:Ofinal3}) simplifies to 
\begin{equation}
    \begin{aligned}
        \nabla_{\tilde{\bm{\rho}}_t}\log p(\tilde{\bm{\rho}}_t)=\frac{1}{\sigma^2_t}
        \int{\log{(\bm{x} + \epsilon)}}{}&p(\bm{x}|\tilde{\bm{\rho}}_t)d\bm{x} - \frac{\tilde{\bm{\rho}}_t}{\sigma^2_t} 
        % \\&\EX\left[\frac{ {\log{(\bm{x} + \epsilon)}}-\tilde{\bm{\rho}}_t}{\sigma^2_t}\bigg|\tilde{\bm{\rho}}_t\right]
        \;.
    \label{eq:gradrho}
    \end{aligned}
\end{equation}

Define by
$\EX_{\bm{x}\sim p(\bm{x}|\tilde{\bm{\rho}}_t)}$ the operation of expectation over all objects $\bm{x}$, when $\bm{x}$ is sampled 
according to the PDF
$p(\bm{x}|\tilde{\bm{\rho}}_t)$. 
%given $\tilde{\bm{\rho}_t}$.
Then, 
\begin{equation}
     \int{\log{(\bm{x} + \epsilon)}}p(\bm{x}|\tilde{\bm{\rho}}_t)d\bm{x}=
    \EX_{\bm{x}\sim p(\bm{x}|\tilde{\bm{\rho}}_t)}\left[{ {\log{(\bm{x} + \epsilon)}}}\right] \;.
  \label{eq:intE}
\end{equation}
Therefore, Eq.~(\ref{eq:gradrho}) can be written as
\begin{equation}
    \nabla_{\tilde{\bm{\rho}_t}}\log p(\tilde{\bm{\rho}}_t) = \frac{\EX_{\bm{x}\sim p(\bm{x}|\tilde{\bm{\rho}}_t)}\left[{ {\log{(\bm{x} + \epsilon)}}}\right] - \tilde{\bm{\rho}}_t}{\sigma^2_t}\;.
\label{eq:Ofinal5}
\end{equation}
% Equivalently, using the definitions in Eq.~(\ref{eq:x_rho}, \ref{eq:Ofinal5}),
% \begin{equation}
%     \nabla_{\tilde{\bm{\rho}_t}}\log p(\tilde{\bm{\rho}}_t) = \frac{\EX_{\bm{\rho}_c\sim p(\bm{\rho}_c|\tilde{\bm{\rho}}_t)}\left[{ {\bm{\rho}_c}}\right] - \tilde{\bm{\rho}}_t}{\sigma^2_t}\;.
% \label{eq:Ofinal5_rho}
% \end{equation}
We do not hold the PDF $p(\bm{x}|\tilde{\bm{\rho}}_t)$, partly because we do not have explicit knowledge on how nature is distributed. To address this problem, the term ${\EX_{\bm{x}\sim p(\bm{x}|\tilde{\bm{\rho}}_t)}}\left[{ {\log{(\bm{x} + \epsilon)}}}\right]$ is approximated via a state of the art DNN. The expectation term of a form 
${\EX_{\bm{x}\sim p(\bm{x}|\tilde{\bm{\rho}}_t)}}\left[\cdot \right]$ has been proved in~\cite{vincent2011connection} to be equivalent to the output of an ideal \textit{denoiser}. In practice, a trained denoising DNN approximates this term. 

%Denoising score matching~\cite{vincent2011connection}, proved the objective (i.e. ${\EX_{\bm{x}\sim p(\bm{x}|\tilde{\bm{\rho}}_t)}}\left[{ {\log{(\bm{x} + \epsilon)}}}\right]$) to be equivalent to the output of a trained \textit{Denoiser}. 

% Because ${\EX_{\bm{x}\sim p(\bm{x}|\tilde{\bm{\rho}}_t)}}\left[{ {\log{(\bm{x} + \epsilon)}}}\right]$ averages over $\bm{x}$, the expectation result is only a function of $\tilde{\bm{\rho}}_t$. So, a DNN that estimates this expectation operates on
% $\tilde{\bm{\rho}}_t$. 

\begin{figure*}[t]
    \centering
    \includegraphics[width=1\textwidth]{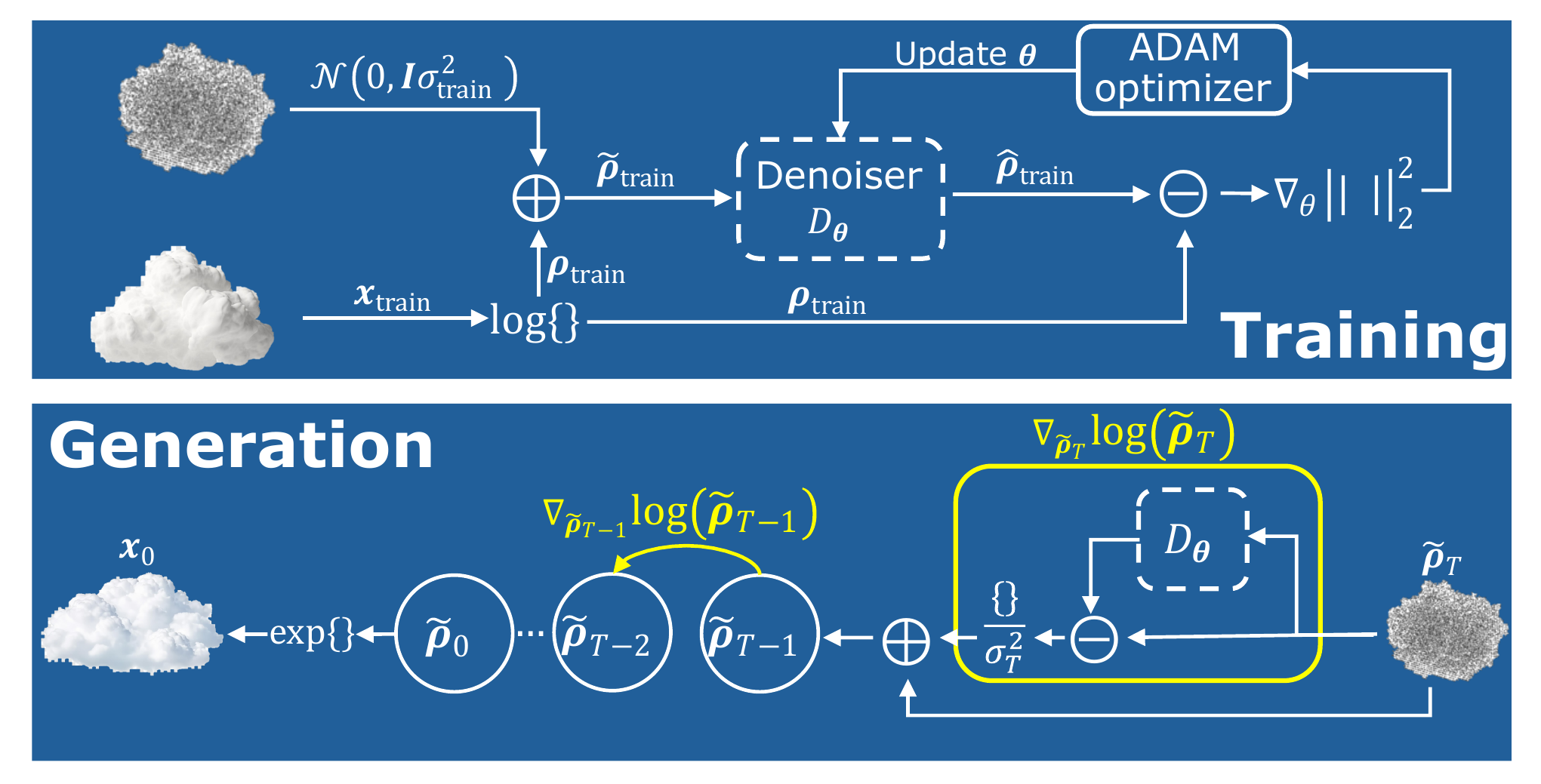}
    \caption{Flowcharts: the DNN training process (Algorithm~1) and the generative nonnegative diffusion (NnD) process (Algorithm~2) yielding a nonnegative object.}
    \label{fig:flowchart}
\end{figure*}

The denoiser is a DNN having a parameter set $\bm{\theta}$, that takes a noisy object as input and produces a clean object as output. The denoiser trains on labeled examples: each example has a true object $\bm{x}_{\rm train}$ and corresponding corrupted latent fields, denoted $\bm{\rho}_{\rm train}$. This allows the denoiser to implicitly learn the expectation.
A set of true objects $\bm{x}_{\rm train}$ is obtained by an external database of natural non-negative objects of our type of interest.  Per $\bm{x}_{\rm train}$, we create a variety of synthetic latent fields, in analogy to  Eq.~(\ref{eq:tilderhot}):
\begin{equation}
       {\bm{\rho}_{\rm train}} = 
       \log{({\bm{{x}}_{\rm train}} +\epsilon)} + \mathcal{N}(0, \sigma^2_{\rm train})
    \;.
    \label{eq:rhotag}
\end{equation}
Per noise variance $\sigma^2_{\rm train}$, multiple samples $\mathcal{N}(0, \sigma^2_{\rm train})$ are generated. Moreover, multiple values of $\sigma^2_{\rm train}$ are used. 
%corrupted with different noise standard deviations are created synthetically, 
Supervised learning trains the DNN by minimizing a loss
\begin{equation}
    {\bm{\theta}} = 
    \arg\min_{\bm{\theta}} 
    \| 
    \log(\bm{x}_{\rm train} + \epsilon) 
     - D_{\bm{\theta}}({\bm{\rho}_{\rm train}})
     \|^2_2 \;. 
    \label{eq:DNN_complex}
\end{equation}
This process is summarized in Algorithm~1 and Fig.~\ref{fig:flowchart}.

After training, 
\begin{equation}
\label{eq:approx}
    D_{\bm{\theta}}(\tilde{{\bm{\rho}_t}})\approx {\EX_{\bm{x}\sim p(\bm{x}|\tilde{\bm{\rho}}_t)}}\left[{ {\log{(\bm{x} + \epsilon)}}}\right]
    \;.
\end{equation}
The output of the DNN is
\begin{equation}
\label{eq:hatrho}
    \hat{\bm{\rho}}_t=D_{\bm{\theta}}({{\tilde{\bm{\rho}}_t}})
    \;.
\end{equation}
Using Eqs.~(\ref{eq:Ofinal5},\ref{eq:approx}), the score network result $s_{\bm{\theta}}(\tilde{\bm{\rho}_t})$ is
\begin{equation}
  \nabla_{\tilde{\bm{\rho}_t}}
  \log p(\tilde{\bm{\rho}}_t) 
  \approx 
   s_{\bm{\theta}}(\tilde{\bm{\rho}_t})  =  \frac{D_{\bm{\theta}}({{\tilde{\bm{\rho}}_t}}) - \tilde{\bm{\rho}}_t}{\sigma^2_t}\;.
\end{equation}
Overall, the generation process is summarized in Algorithm~2
and Fig.~\ref{fig:flowchart}.
\begin{algorithm}[h]
    \caption*{\textbf{Algorithm 1:} Training}
    \begin{algorithmic}[1]
        \Require $ L, {\epsilon}, \{\sigma_{\rm train}^{(i)}\}^L_{i=1} 
            $\\
            ${\textbf{initialize}~\bm{\theta}}$
            \Repeat
                \State $\bm{x}_{\rm train} \sim {\rm Sample~from ~database}$
                \State $i \sim \text{Uniform}(\{1, \dots, L\})$
                \State $\bm{n}_{\rm train} \sim 
                \mathcal{N}
                \left( 0, {\bm I}[\sigma^{(i)}_{\rm train}]^2
                \right)$
                \State $\bm{\rho}_{\rm train} = 
                \log{(\bm{x}_{\rm train} + \epsilon}) + \bm{n}_{\rm train}$
                \State Take gradient descent step using
                \Statex \hspace{1cm} $\nabla_{\theta} \left\| \log{(\bm{x}_{\rm train} + \epsilon)} - D_{\theta} \left(\bm{\rho}_{\rm train}\right) \right\|^2_2$
            \Until{convergence} \State
            \Return $\bm{\theta}$
    \end{algorithmic}
\end{algorithm}

%$s_{\bm{\theta}}(\tilde{\bm{\rho}_t})$ is
%estimated by,
% \begin{equation}
%    s_{\bm{\theta}}(\tilde{\bm{\rho}_t})  = \nabla_{\tilde{\bm{\rho}_t}}\log p(\tilde{\bm{\rho}}_t) = \nabla_{\tilde{\bm{\rho}_t}}\log p(\tilde{\bm{\rho}}_t) = \frac{D_\theta({{\tilde{\bm{\rho}}_t}}) - \tilde{\bm{\rho}}_t}{\sigma^2_t}\;.
% \end{equation}
% This approach is only viable when 3D data is available, which is essential for many 3D reconstruction tasks, as discussed in~\cref{sec:intro}. To address this limitation, we used 3D simulated data (from the Weizmann Institute). However, a challenge with simulated 3D data is that it often does not reflect the statistical properties of empirical data from the real world. As a result, the learned probability distribution, $\nabla_{\bm{x}_t}\log{p(\bm{x}_t)}$, may not accurately represent the true distribution. To balance this, we propose a loss function that incorporates both simulated and empirical data.
% \subsubsection{Training based on 2D data}
% \subsubsection{Training based on 3D data}

%%%%%%%%%%%%%%%%%%%%%%%%%%%%%%%%%%%
\subsection{Gradient of the Data Term}
\label{sec:graddata}

% % We now describe how $\nabla_{\tilde{\bm{\rho}}_t}\log{p_{\tilde{\bm{\rho}}}(\bm{y}|\tilde{\bm{\rho}}_t)}$ is estimated.

% The term $p(\bm{y}|{{\bm{{x}}}})$ can be analytically traced. Given the forward model,  and the true physical object $\bm{x}$, the posterior distribution $p(\bm{y}|{{\bm{{x}}}})$ is simply defined by the noise PDF,   
% \begin{equation}
% \begin{aligned}
%     p(\bm{y}|{{\bm{{x}}}}) &= p(\bm{n}=\bm{y} - \mathcal{F}(\bm{x})|{{\bm{{x}}}}) \\& = \frac{1}{\sqrt{2\pi \bm{y}}}\exp\left[{-\frac{(\bm{y} - \mathcal{F}({\bm{x}}))^2}{2\bm{y}}}\right].
% \end{aligned}
% \end{equation}
% Then the gradient follows,
% \begin{equation}
% \label{eq:grad_p_yx}
% \nabla_{{\bm{x}}}\log p(\bm{y}|{{\bm{{x}}}}) = \frac{\bm{y} - \mathcal{F}({\bm{x}})}{\bm{y}}\frac{\partial F(\bm{x})}{\partial x}
% \end{equation}
% As detailed in~\cref{sec:nonngeg}, 

We apply diffusion steps in the latent-space $\bm{\rho}$. On the other hand, a forward model expects to run on physically feasible objects $\bm{x}$. We thus account for these two aspects.

%adjustment to the forward model must be made. We imagine the forward model as there was an $\exp\{\cdot\}$ at the input. The added $\exp\{\cdot\}$ can be simply inversed via $\log\{\cdot\}$, hence, no information is lost. 
% Given $\bm{\rho}$, the measurements $\bm{y}$ are captured through,
% \begin{equation}
%     \bm{y} = \mathcal{F}\{\exp\bm{[x]}\} + \mathcal{N}(0,\bm{y})
% \end{equation}
% Analogically to~\cref{eq:grad_p_yx}, 
The posterior PDF $p(\bm{y}|{{\bm{{\rho}}}})$  is defined through the imaging noise PDF. Following
Eq.~(\ref{eq:p_y_xk}), 
\begin{equation}
\label{eq:p_y_rho}
\begin{aligned}
      p(\bm{y}|{{\bm{{\rho}}}}) 
      %&= p(\bm{n}|{{\bm{{\rho}}}}) =
      %p(\bm{y} - \mathcal{F}\{\exp[\bm{\rho}]\}|%{{\bm{{\rho}}}})
      %\\& = 
      \approx \frac{1}{\sqrt{2\pi \bm{y}}}\exp\left[{-\frac{(\bm{y} - \mathcal{F}\{{\exp\left[\bm{\rho}\right]}\})^2}{2\bm{y}}}\right].
\end{aligned}
\end{equation}
% Therefore, the gradient of the posterior PDF is  
% \begin{equation}
%     \begin{aligned}
%     \label{eq:grad_p_yrho}
%     \nabla_{\bm{{\rho}}}\log{}&{p(\bm{y}|{{\bm{{\rho}}}})} = \frac{\bm{y} - \mathcal{F}\{\exp\left[{\bm{{\rho}}}\right]\}}{\bm{y}}
%     \frac{\partial F\{\exp\left[{\bm{{\rho}}}\right]\}}{\partial \exp\left[{\bm{{\rho}}}\right]}\frac{\partial\exp\left[{\bm{{\rho}}}\right]}{\partial\bm{\rho}}
%     \;.
%     \end{aligned}
% \end{equation}

Contrary to Eq.~(\ref{eq:p_y_rho}), which depends on a true 
$\bm{\rho}$, our ALD takes diffusion steps in the latent space 
$\tilde{\bm{{\rho}}}_t$, using $\nabla_{\tilde{\bm{\rho}_t}}\log p(\bm{y}|{\tilde{\bm{{\rho}}}}_t)$. %Unlike~(\ref{eq:grad_p_yrho}), 
The term $p(\bm{y}|{\tilde{\bm{{\rho}}}}_t)$ is affected by two noise contributions: those of the measurement and annealing. Disentangling their PDFs is not straightforward, challenging analytic tracing of $p(\bm{y}|{\tilde{\bm{{\rho}}}}_t)$. To address this challenge, prior art~\cite{torem2023complex,kawar2021snips,kawar2021stochastic} % wisely 
%defined the relation between these two noises. The key innovation in %\cite{torem2023complex, kawar2021snips, kawar2021stochastic}, lies in

\begin{algorithm}[h]
    \caption*{\textbf{Algorithm 2:}~ Generation}
    \begin{algorithmic}[1]
        \Require  $T, \epsilon,\{\sigma_t\}^T_{t=1}, 
        \{\alpha_t\}^T_{t=1}$
            \State Initialize $\tilde{\bm{\rho}}_T\sim\mathcal{N} (\log {\epsilon, \sigma^2_{T}}) ~{\rm or} $ $\mathcal{N} (0, \sigma^2_{T})$
            \For{$t=T,\dots,1$} 
 %               \For{{$t=k,\dots,1$} }
                    \State $\bm{\eta}_t\sim\mathcal{N}(0, \bm{I})$
                     % \State  $\tilde{\bm{\rho}}_{t-1} = \tilde{\bm{\rho}}_t + \alpha_t D_{\theta}(\tilde{\bm{\rho}}_t) +\sqrt{2\alpha_t}\bm{\eta}_t $
    %                \State $ \nabla_{\tilde{\bm{\rho}}_t}\log p(\tilde{\bm{\rho}}_t) = 
     %     \frac{D_{\theta}(\tilde{\bm{\rho}}_t) - \tilde{\bm{\rho}}_t}{\sigma^2_t}$
                    \State  $\tilde{\bm{\rho}}_{t-1} = 
                    \tilde{\bm{\rho}}_t + \alpha_t
                    \frac{D_{\theta}(\tilde{\bm{\rho}}_t) - \tilde{\bm{\rho}}_t}
                         {\sigma^2_t}
                    +\sqrt{2\alpha_t}\bm{\eta}_t $
 %               \EndFor
            \EndFor
            \State $\bm{x}_0 = \exp\left[\tilde{\bm \rho}_0\right]$
            \State \Return $\bm{x}_0$
    \end{algorithmic}
\end{algorithm}

innovated ways to eliminate statistical dependencies between the annealing and measurement noise components in the posterior PDF. However, such mathematical maneuvers become excessively complex as a forward imaging model  becomes more complex. We refer here to complexity raising, for example, in recursive non-linear models such as 3D radiative transfer, or stochastic models as timing of single-photon detection~\cite{lindell2018single, ma2023burst}.

We propose to make use of a different mathematical approach, introduced in~\cite{chung2022diffusion}. This approach does not analytically track
$p(\bm{y}|{\tilde{\bm{{\rho}}}}_t)$, but approximates it. Recall that 
$\bm{\rho}$ is a noiseless representation of an object in our latent space. The PDF $p(\bm{y}|{\tilde{\bm{{\rho}}}}_t)$ can be defined by 
%a convolution of two others,
\begin{equation}
  \begin{aligned}
    p(\bm{y}|{\tilde{\bm{{\rho}}}}_t) = %\int{p(\bm{y}|\bm{\rho}_{0},{\tilde{\bm{{\rho}}}}_t)p(\bm{\rho}_{0}|{\tilde{\bm{{\rho}}}}_t)d\rho_0}= \\
    \int{p(\bm{y}|\bm{\rho})p(\bm{\rho}|{\tilde{\bm{{\rho}}}}_t)d\bm{\rho}}=\EX_{\bm{\rho}\sim p(\bm{\rho}|{\tilde{\bm{{\rho}}}}_t)}\left[p(\bm{y}|\bm{\rho})\right]
    \;.
    \label{eq:conv}
    \end{aligned}
\end{equation}
The expectation operator in Eq.~(\ref{eq:conv}) bears resemblance to 
the operator in Eqs.~(\ref{eq:intE},\ref{eq:Ofinal5}), with a similar limitation and part of a practical solution. 
We do not hold the PDF $p(\bm{\rho}|{\tilde{\bm{{\rho}}}}_t)$, partly because we do not have explicit knowledge on how nature (equivalent to $\bm{\rho}$) is distributed.

Following~\cite{chung2022diffusion}, let us approximate\footnote{This approximation is closely related to Jensen’s inequality~\cite{simic2008global}.} the PDF (\ref{eq:conv}) by
\begin{equation}
  \begin{aligned}
    p(\bm{y}|{\tilde{\bm{{\rho}}}}_t) \approx %\int{p(\bm{y}|\bm{\rho}_{0},{\tilde{\bm{{\rho}}}}_t)p(\bm{\rho}_{0}|{\tilde{\bm{{\rho}}}}_t)d\rho_0}= \\
   %p(\bm{y} |\int{p(\bm{\rho}|{\tilde{\bm{{\rho}}}}_t)d\bm{\rho}})
   p
   \left(
     \bm{y}|~\EX_{\bm{\rho}\sim p(
     \bm{\rho}|{\tilde{\bm{{\rho}}}}_t)}\left[\bm{\rho}\right]
   \right).
       \label{eq:approxPDF}
    \end{aligned}
\end{equation}
In other words, the PDF is assessed given an expectation based on 
$\tilde{\bm{\rho}}_t$, rather than directly on $\tilde{\bm{\rho}}_t$.
The expectation operator ${\EX_{\bm{x}\sim p(\bm{x}|\tilde{\bm{\rho}}_t)}}\left[{ {\log{(\bm{x} + \epsilon)}}}\right]$ appearing in Eq.~(\ref{eq:approx}) is equivalent to $\EX_{\bm{\rho}\sim p(\bm{\rho}|{\tilde{\bm{{\rho}}}}_t)}\left[\bm{\rho}\right]$ appearing in 
Eq.~(\ref{eq:approxPDF}). So, as discussed in~\cref{sec:gradprior}, this operator can be estimated using a trained denoiser ${D_{\bm{\theta}}({{\tilde{\bm{\rho}}_t}})}$.
%can be used as an estimator for ${\EX_{\bm{x}\sim p(\bm{x}|\tilde{\bm{\rho}}_t)}}\left[{ {\log{(\bm{x} + \epsilon)}}}\right]$, which is equivalent to $\EX_{\bm{\rho}\sim p(\bm{\rho}|{\tilde{\bm{{\rho}}}}_t)}\left[\bm{\rho}\right]$. 

% \begin{equation}
%   \begin{aligned}
%     p(\bm{y}|{\tilde{\bm{{\rho}}}}_t) \approx %\int{p(\bm{y}|\bm{\rho}_{0},{\tilde{\bm{{\rho}}}}_t)p(\bm{\rho}_{0}|{\tilde{\bm{{\rho}}}}_t)d\rho_0}= \\
%     p(\bm{y} |\int{p(\bm{\rho}|{\tilde{\bm{{\rho}}}}_t)d\bm{\rho}})=p(\bm{y}|\EX_{\bm{\rho}\sim p(\bm{\rho}|{\tilde{\bm{{\rho}}}}_t)}\left[\bm{\rho}\right])
%     \label{eq:approxPDF}
%     \end{aligned}
% \end{equation}

% To address this problem, the term ${\EX_{\bm{x}\sim p(\bm{x}|\tilde{\bm{\rho}}_t)}}\left[{ {\log{(\bm{x} + \epsilon)}}}\right]$ is approximated via a state of the art DNN. The expectation term of a form 
%${\EX_{\bm{x}\sim p(\bm{x}|\tilde{\bm{\rho}}_t)}}\left[\cdot \right]$ has been proved in~\cite{vincent2011connection} to be equivalent to the output of an ideal \textit{denoiser}. In practice, a trained denoising DNN approximates this term.

%Unfortunately, we do not have access to the probability distribution $p(\bm{\rho}|{\tilde{\bm{{\rho}}}}_t)$. Hence, the calculation of $\EX_{\bm{\rho}\sim p(\bm{\rho}|{\tilde{\bm{{\rho}}}}_t)}\left[p(\bm{y}|\bm{\rho})\right]$ cannot be done. Instead of calculating the above, ~\cite{chung2022diffusion} approximates\footnote{This approximation is closely related to Jensen’s inequality~\cite{simic2008global}} $p(\bm{y}|{\tilde{\bm{{\rho}}}}_t)$ by,

The denoiser (\ref{eq:hatrho}) outputs an estimated clean field $\hat{\bm{\rho}}_t$.
%=D_{\bm{\theta}}({{\tilde{\bm{\rho}}_t}})$. 
Then, following Eqs.~(\ref{eq:p_y_rho},\ref{eq:approxPDF}), 
\begin{equation}
    \label{eq:tildep_y_rho}
    \begin{aligned}
     p(\bm{y}|{\tilde{\bm{{\rho}}}}_t) & \approx
      p(\bm{y}|\hat{\bm{\rho}}_t) 
            = \frac{1}{\sqrt{2\pi \bm{y}}}\exp
           \left[
             {-\frac{(\bm{y} - \mathcal{F}
             \{
               {\exp\left[\hat{\bm{\rho}}_t \right]}
             \})^2}{2\bm{y}}}\right].
    \end{aligned}
\end{equation}
Consequently, %following Eq.~(\ref{eq:grad_p_yrho}),
\begin{equation}
    \label{eq:grad_practical}
    \begin{aligned} 
             \nabla_{\tilde{{\bm{{\rho}}}}_t}\log{}&{p(\bm{y}|{\tilde{{\bm{{\rho}}}}}_t)} \approx\\& \frac{\bm{y} -  \mathcal{F}\{{\exp[\hat{\bm{\rho}}_t]}\}}
             {\bm{y}}\frac
             {\partial \mathcal{F}
             \{\exp\left[\hat{\bm{\rho}}_t\right]\}}
             {\partial \exp\left[\hat{\bm{\rho}}_t\right]}
             \frac{\partial\exp\left[\hat{\bm{\rho}}_t\right]}{\partial\hat{\bm{\rho}}_t}
             \frac{\partial \hat{\bm{\rho}}_t}{\partial{\tilde{\bm{{\rho}}}}_t}
             \;.
          \end{aligned}
\end{equation}
Eq.~(\ref{eq:grad_practical}) includes the term  
$\frac{\partial{ \hat{\bm{\rho}}}_t }
      {\partial{\tilde{\bm{{\rho}}}}_t}$.
This term is numerically obtained by backpropagation through the DNN that implements Eq.~(\ref{eq:hatrho}).

\section{Examples}
\label{sec:exper}

In this work, $\bm{x}$ is a 3D {\em volumetric} cloud scene. Here, in each voxel there is a nonnegative vector having three elements. Each vector element represents a microphysical parameter: the liquid water content (LWC), the effective radius $r_{\rm e}$ and effective variance $v_{\rm e}$. We first describe some technical implementation details. Then, we show results.

%%%%%%%%%%%%%%%%%%%%%%%%%%%%% 
\subsection{Implementation}
\label{sec:implement}

For numerical stability, we use  $\epsilon = \exp(-10)$ in the latent representation. Furthermore,
the different microphysical parameter tend to be quoted in units where their values have very different orders of magnitude. To better condition the training (Algorithm~1), we pre-scale the values to have similar orders of magnitude. Then, after scene generation (Algorithm 2), the values are rescaled to their proper units.

In this work, ${\bm{{x}}_{\rm train}}$ is a simulated 3D cloud scene. Each training scene has a domain of size $32\times 32\times 32$ voxels, being $1.6\times 1.6{\rm km}^2$ wide and $1.28{\rm km}$ thick. These training scenes were cropped from larger cloud fields that exist in online datasets~\cite{ronen20253d}. These large cloud fields had been generated using physics-based dynamical simulations, based on boundary conditions termed BOMEX~\cite{heiblum2019core}. There are 23,040 examples of ${\bm{{x}}_{\rm train}}$  in the training database~\cite{ronen_2025}.

The DNN trains according to Algorithm~1. Here, we follow Ref.~\cite{song2020improved}: there are $L$ values for the training noise standard deviation, defined by a geometric sequence
%$     \sigma_{\rm train}^{(i)} =      \sigma_{\rm train}^{(L)} r^{i} 
%$, 
%where $i=1\ldots L$ and 
\begin{equation}
\label{eq:geo_seq_r}
 %   r = \sqrt[L]{\frac{\sigma_{\rm train}^{(0)}}{\sigma_{\rm train}^{(L)}}}
 %   r = %\sqrt[L]{\frac
    \sigma_{\rm train}^{(i)} =  \sigma_{\rm train}^{(L)}
     \left[
      \sigma_{\rm train}^{(1)}/\sigma_{\rm train}^{(L)}
     \right]^{(L-i)/(L-1)}   % {i/L} 
         ,~
       {\rm where}~~ i=1\ldots L
       .
\end{equation}
%Then, 
%\begin{equation}
%    \label{eq:sigmai}
%      \sigma_{\rm train}^{(i)} = 
%      \sigma_{\rm train}^{(L)}\cdot r^{i}
%       \;
%       ~{\rm where}~~~ i=1\ldots L
%       \;.
% \end{equation}
In each training iteration, the index $i$ is drawn randomly from a uniform distribution. 
We used $L=150$, $\sigma_{\rm train}^{(1)}=10^{-2}$ and 
${\sigma_{\rm train}^{(L)}=10^{2}}$. 
The DNN has a 3D U-Net architecture~\cite{cciccek20163d}, which suits volumetric voxel grids. We use a DNN implementation from~\cite{10.7554/eLife.57613}. 
Training was conducted on batches of 32 example scenes. The ADAM optimizer~\cite{kingma2014adam} was used with a learning rate of $10^{-4}$ and a weight decay of $10^{-5}$.
Training was performed for approximately 50,000 
steps over the course of $\sim5$ hours on an NVIDIA GeForce RTX 3090 GPU.

After DNN training, Algorithm~2 generates 3D cloud scenes. Here $T=600$.
%The values $\sigma_t$ and $\alpha_t$ change every $K=5$ iterations. 
%In analogy to Eq.~(\ref{eq:geo_seq_r}), 
The parameter representing the annealing noise standard deviation changes every $K=5$ iterations and follows a staircase function,
\begin{equation}
         \sigma_t = 
      \sigma_{1} 
      \left[
         \sigma_T/\sigma_1
      \right]^
  %    {{({T/5} - \lfloor{(t-1)/5}\rfloor })/{({T/5})}} 
  %    {{({T/5} -1 - \lfloor{(t-1)/5}\rfloor })/{({T/5} -1)}} 
%     {{(T -5 - 5\lfloor{(t-1)/5}\rfloor })/(T-5)} 
%     {{ [5/(T-5)]\lfloor{(t-1)/5}\rfloor }} 
    {{ [K/(T-K)]\lfloor{(t-1)/K}\rfloor }} 
%       \sigma_{T} 
%       \left[
%          \sigma_1/\sigma_T
%       \right]^
%   %    {{({T/5} - \lfloor{(t-1)/5}\rfloor })/{({T/5})}} 
%   %    {{({T/5} -1 - \lfloor{(t-1)/5}\rfloor })/{({T/5} -1)}} 
% %     {{(T -5 - 5\lfloor{(t-1)/5}\rfloor })/(T-5)} 
%      {{1 - [5/(T-5)]\lfloor{(t-1)/5}\rfloor }} 
     \;
       ~~{\rm where}~~~ t=1\ldots T
       \;.
       \label{eq:sigmat}
\end{equation}
Here, $\sigma_T=10^2$ and $\sigma_1=10^{-2}$. 
Then,
%\begin{equation}
$         \alpha_t=
         \zeta [\sigma_t/ \sigma_1]^2$,
%         \;.
%       \label{eq:alphat}
%\end{equation}
where $\zeta=2\cdot 10^{-6}$. Generation was done on NVIDIA GeForce RTX 3090 GPU. Generating each scene took $\sim 2[s]$.
Our code and model will be placed in the public domain upon publication of this paper. 
% \begin{equation}
%        r = \sqrt[\lfloor{T/5}\rfloor]{\frac{\sigma_{0}}{\sigma_{\lfloor{T/5}\rfloor}}}
 % \end{equation}
% Then, 
% \begin{equation}
%          \sigma_t = 
%       \sigma_{T}\cdot r^{\lfloor{T/5}\rfloor}
%        \;
%        ~{\rm where}~~~ t=1\ldots T
%        \;.
% \end{equation}

% \begin{equation}
%     \label{eq:sigmai}
%       \sigma_{t}=
%       \left[
%       \frac{\sigma_{(T)}}
%            {\sigma_{(1)}}
%        \right]^{\lfloor{t/5}\rfloor},~~~\alpha_t = \sigma_t
%        \;
%        ~{\rm where}~~~ t=1\ldots T
%        \;.
%  \end{equation}
%  according to
% $\sigma_t = \lfloor{t/5}\rfloor$

% = 100, "sigma_1" = 0.01, T = 120, k = 5, 
%  "alpha t" = 2e-6* (sigma t/sigma1) . 
% The geometric progression is the same as we use for the noise scales in the training of the MMSE. "sigma 0" is set to 100 and n_step_each is set to 120 in the generation 

% from random noise. We show the results in Fig 1, 2.  A

%%%%%%%%%%%%%%%%%%%%%%%%%%%%% 
\subsection{Results}
\label{sec:result}

\begin{figure}[t]
    \centering
    \includegraphics[width=0.5\textwidth]{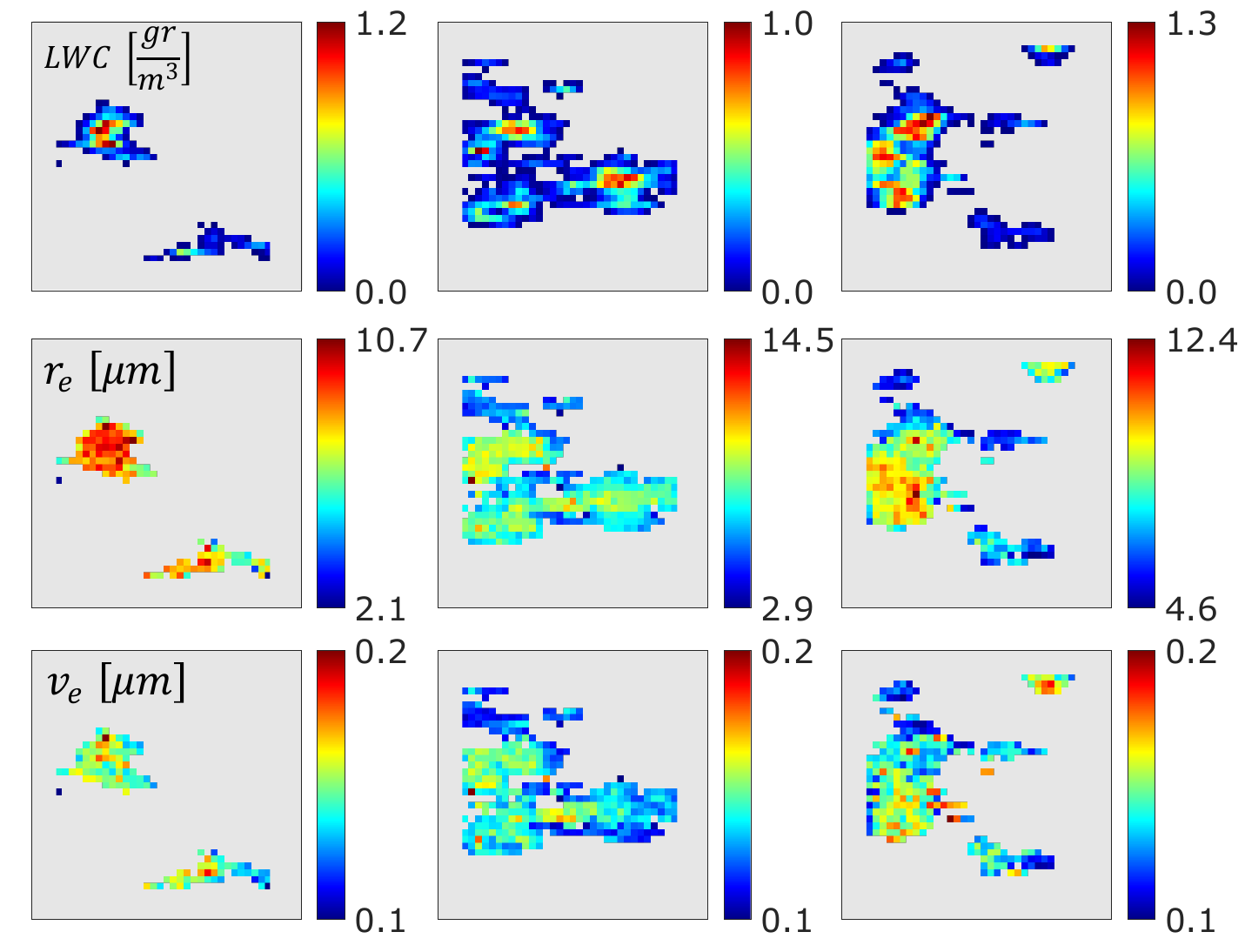}
     \caption{Maximum intensity projection (MIP) 
     onto the horizontal plane, along the vertical axis.
     Each row corresponds to a different generated 3D cloud scene in a horizontal $1.6\times 1.6[{\rm km}^2]$ segment.
     [Top] Liquid Water Content, ${\rm LWC}$. [Middle] Effective radius, $r_{\rm e}$. [Bottom] Effective variance, $v_{\rm e}$.}
    \label{fig:XYgen}
\end{figure}

As mentioned in Sec.~\ref{sec:implement}, training used domains that are horizontally cropped out of large cloud fields. Consequently, some training examples have an entire cloud or several clouds wholly contained in the horizontal domain, but sometimes clouds are cropped at a domain boundary. This distribution is also seen in examples of generated clouds, displayed at top-view using maximum intensity projection (MIP) in 
Fig.~\ref{fig:XYgen}. 

\begin{figure}[t]
    \centering
    \includegraphics[width=0.5\textwidth]{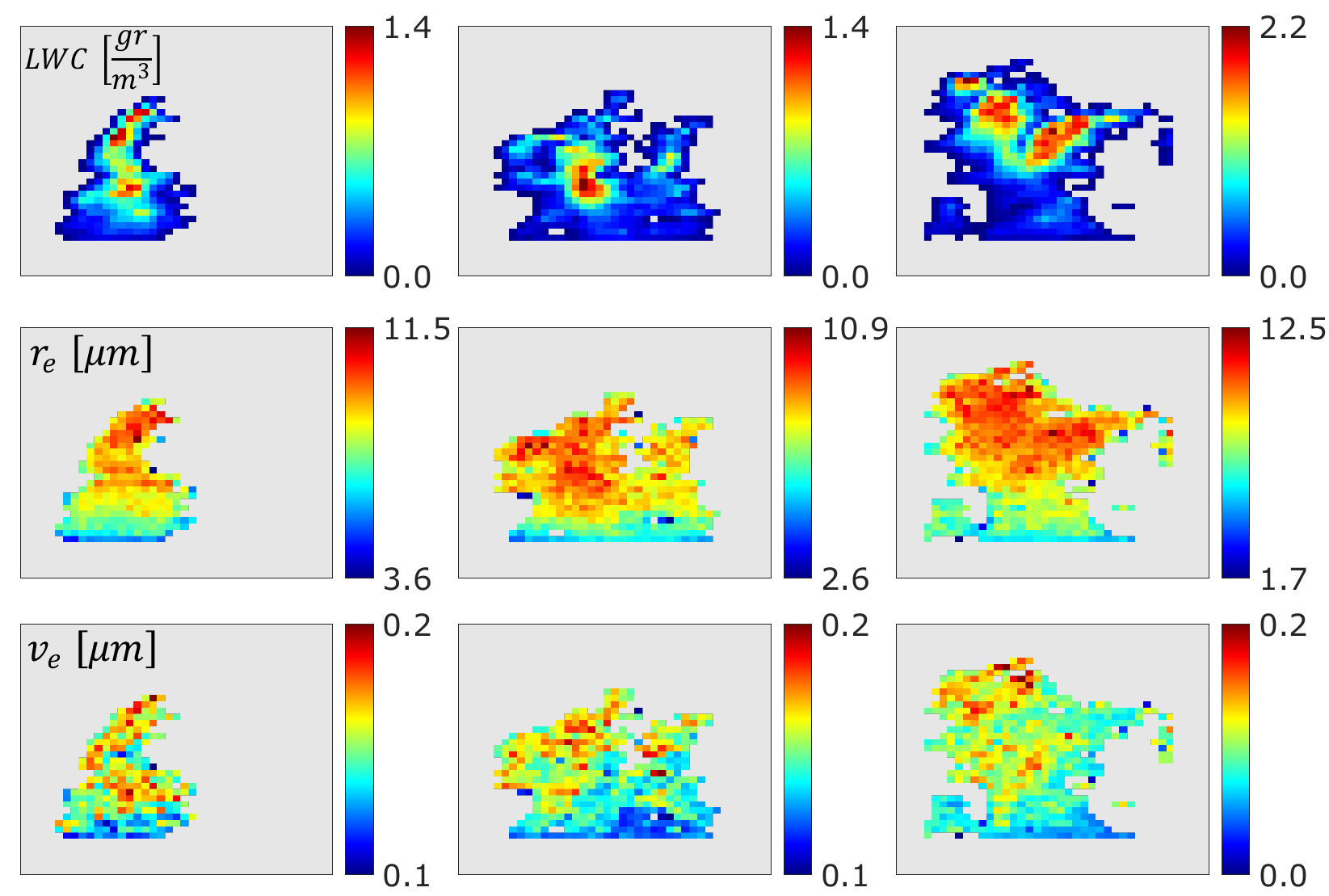}
    \caption{MIP to a side view of generated 3D cloud scenes. Each projection is 
    $1.6{\rm km}$ wide and has height range of $1.28{\rm km}$. 
         [Top] Liquid Water Content, ${\rm LWC}$. [Middle] Effective radius, $r_{\rm e}$. [Bottom] Effective variance, $v_{\rm e}$.
    }
    \label{fig:XZgen}
\end{figure}

Side-views of some generated clouds are shown in Fig.~\ref{fig:XZgen}. 
These are interesting. They are consistent with trends described in~\cite{betzer2024nemf} regarding cloud cores, away from the cloud top and lateral periphery. There, evaporation and mixing with surrounding air are minimal, leading to rather consistent trends  of increasing LWC and $r_{\rm e}$ with altitude. Moreover, $v_{\rm e}$ has a trend of generally increasing with $r_{\rm e}$, in consistency with~\cite{zhang2011vertical}.

\begin{figure}[t]
    \centering
    \includegraphics[width=0.48\textwidth]{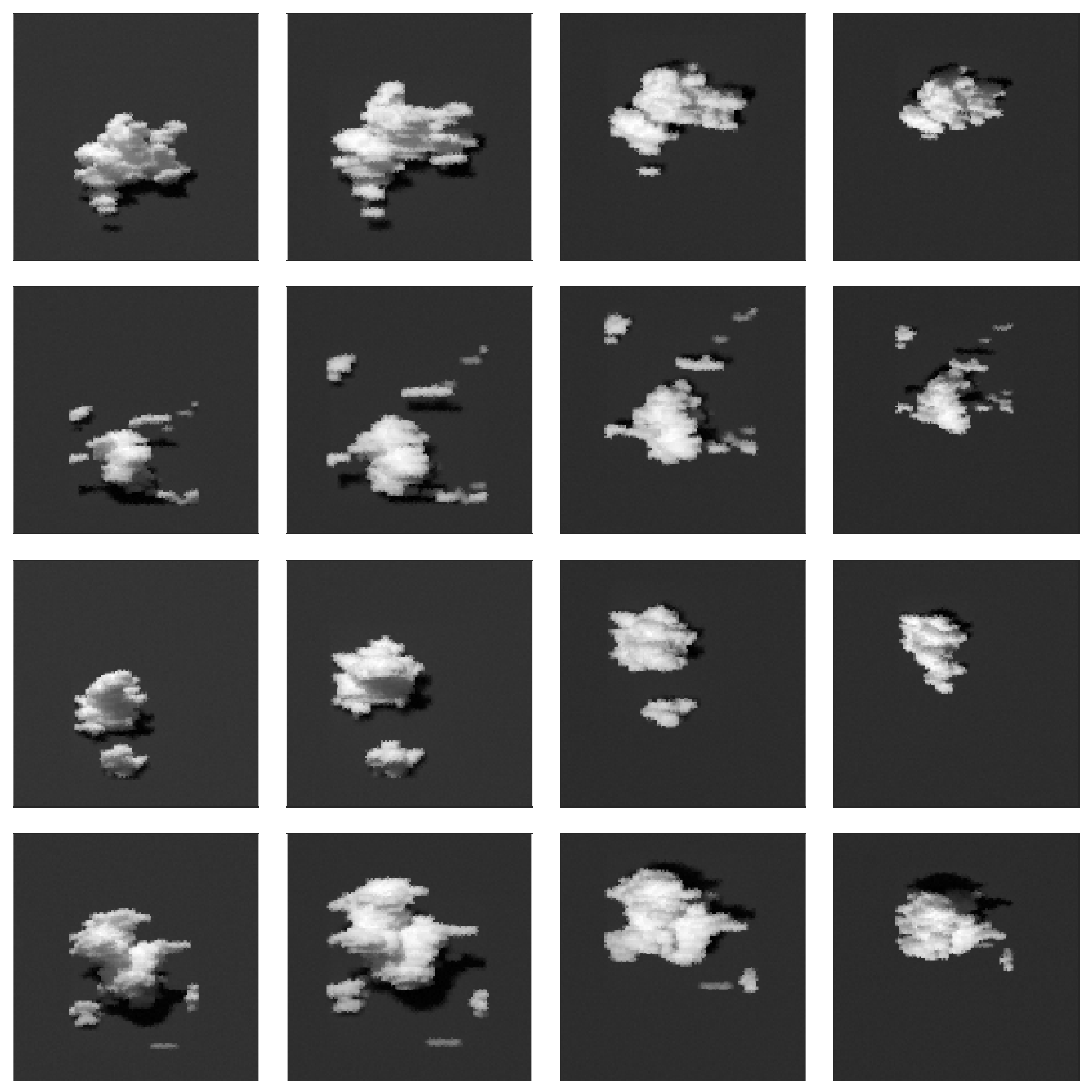}
    \caption{Intensity images of clouds, 
    rendered based on our synthetically generated 3D cloud structures. Rendering integrates radiance in the range  $500-600{\rm nm}$. The sun is at azimuth $46^\circ$ and elevation $71^\circ$.  There is a distinct cloud scene per row.  Each column has a distinct viewing direction (off nadir). From left to right: 
    $[{\rm Azimuth}, {\rm Off~nadir}]=$
    $[0^\circ,77.6^\circ]$, $[0^\circ,86.9^\circ]$,
    $[180^\circ,83.81^\circ]$ and $[180^\circ,74.5^\circ]$, respectively.
    For display clarity, all images are gamma-corrected.}
    % 0,3,6,9 (from 0 to 9 )
    \label{fig:cloudimages}
\end{figure}

The forward model was applied to some of our generated clouds. This was done using an online public domain code\footnote{https://github.com/inbalkb/NeMF}. This rendered top-view images of clouds over an ocean, taken by a high altitude camera, as seen in Fig.~\ref{fig:cloudimages}.

% \begin{figure*}[t]
%     \centering
%     \includegraphics[width=1\textwidth]{figures/3d_cloud.pdf}
%     \caption{Visualization of the LWC microphysical parameter. We present 6 different generated 3D cloud scenes within a $1.6{\rm km}\times 1.6{\rm km}\times1.28{\rm km}$ volume. The scenes are viewed from Azimuth $-37.5^\circ$ and elevation $30^\circ$}.
%     \label{fig:label_name}
% \end{figure*}

To assess whether our generated 3D clouds are consistent with knowledge about nature, they were judged by an expert, Ilan Koren (Weizmann Institute of Science): a cloud-physics researcher for 35 years. 
%leading an international group and having alumni in senior positions.
His group has 
%They have 
hands-on experience using and developing tools of physical cloud generation.
%and know statistical models of cloud generation. 
%The expert is not a co-author of this paper, unaffiliated with institutions of the co-authors and did not take part in the work.  
The expert was provided with a dataset of 80 volumetric cloud scenes: 40 of them were true samples ${\bm x}_{\rm train}$. The rest were generated by our process. Each scene had three corresponding fields of the microphysical parameters. The dataset scenes were labeled and presented at a random order. 
The {\bf supplementary material} includes the dataset, matlab files to scan scenes, instructions  on how to open the files etc. 

The expert scanned the volumes in any way he liked, and then classified each scene as {\em True} (created by physical simulations) or {\em Fake} (generated by our process). The results as reported to us are summarized in \cref{tab:test_expert}. 
\begin{table}[t]
\centering
\resizebox{0.45\textwidth}{!}{%
    \begin{tabular}{|l|c|c|}
    \hline
    & Classified True & Classified Fake \\
    \hline
    True $\bm{x}_{\text{train}}$ & 35\% & 65\% \\
    \hline
    Generated $\bm{x}_0$ & 57.5\% & 42.5\% \\
    \hline
    \end{tabular}
}
\caption{Classification by a cloud-physics expert, when presented with clouds created by physical simulations and by our generative process. The results seem to indicate that our generated clouds are more consistent with expert perception of clouds.}
\label{tab:test_expert}
\end{table}

The results imply that clouds generated by our process possibly appear more convincing than those created by physical simulations.  The expert knew ahead of time that about half of the clouds are not generated by a physical process. This knowledge possibly biased his classification of many true objects as fake ones. He later noted that it was impossible to detect the fakes. This test implies that our generative process suffices to be consistent with expert perception. This is despite having no human or perceptual elements in the model or training criteria.

%For our experiments, we utilize the 3D U-Net architecture introduced in \cite{cciccek20163d} and implemented in . We use a multi layer U-NET of depth  4, with $[32, 64, 128, 256]$ features, respectively. For training purposes, we use clean (non noisy) simulated 3D cloud scenes \textcolor{red}{[cite Weizman]}. To address the large variations in the values of the microphysical parameters, we normalize each parameter by a known constant. This ensures that the normalized outputs of different microphysical values within the scene are of the same order of magnitude. Using the normalized clean cloud scenes dataset, we generate synthetic pairs of noisy and clean cloud scenes using, as defined in~\cref{eq:defining_o_train}. Finally, the U-Net is utilized as a denoiser. This is done by training on synthetic pairs of clean and noisy normalized scenes, minimizing the L2 norm, as defined in~\cref{eq:DNN_complex}. After training is done, we run a generating task. 

%%%%%%%%%55
\section{Discussion}
\label{sec:discuss}

The paper derives ALD for physically-nonnegative objects.
The approach is especially useful for generating complex scenes that are possible to alternatively compute physically from first principles. The physical models, specifically in flow models, have a huge computational complexity, and this makes the derived ALD significant. It takes {\em days to weeks} to simulate cloud fields from first principles, contrary to {\em seconds} by diffusion. On the other hand,  the feasibility of physical computations enable creation of datasets to  train the diffusion model.

We demonstrated the approach on clouds. We believe it can be useful for generation and analysis of other nonnegative complex object types, listed in Sec.~1.
Moreover, there are additional objects types that have fundamentally nonnegative characteristics, such as\\
$\bullet~$Distance between object elements, e.g., interacting molecules in FRET, adjacent lenses in a barrel.\\
$\bullet~$Lifetime or half-life of object elements, e.g., molecular states in FLIM, nuclear spins in MRI.\\ 
$\bullet~$Energy above a stable equilibrium state (classic) in an object element. In quantum mechanical elements, e.g, vibrating molecules or electromagnetic fields, the energy level is strictly positive above the classic minimum.\\
$\bullet~$Probability. Probability objects are created in light detection. There, radiance of an object element dictates a probability of photon emission, absorption or detection.\\
$\bullet~$Some functions of probability, such as entropy.\\
It may be that such fields can also benefit from this framework.

We used ALD for diffusion. However, in recent years, other diffusion models have been introduced. One notable approach is DDPM~\cite{ho2020denoising}. DDPM is not score-based and does not utilize forward model information. 
%and DDIM~\cite{song2020denoising}. 
%Even so, the framework introduced in~\cite{ho2020denoising}, changes the definition of the noisy object from the ALD definition in~\cref{eq:blurred}. 
Some DDPM principles can be adapted and integrated to the nonnegative diffusion framework that we outline. This may enhance the diffusion performance. 
%An in-depth study of this alternative remains an open direction for future study.

%\textbf{Acknowledgments:} We thank Inbal Kom Betzer for her technical support.  

\ifpeerreview \else
\section*{Acknowledgments}
We are grateful to Ilan Koren, for motivating us to work on this problem, and lending his expertise and time to assess our results experimentally. We thank Inbal Kom Betzer and Vadim Holodovsky for their advice and technical support.

\bibliographystyle{IEEEtran}
% \bibliography{references}

\ifpeerreview \else
%%%% For the camera ready version, please fill out this
%%%% biography. Your camera ready should be within a 12 page limit
%%%% including acknowledgments, references and biography.

% \includegraphics command within an optional argument. (You could
% create your own custom macro containing the \includegraphics command
% to make things simpler here.)
% \begin{IEEEbiography}[{\includegraphics[width=1in,height=1.25in,clip,keepaspectratio]{mshell}}]{Michael Shell}

% \end{IEEEbiography}

%\newpage


%\vfill

\fi

\end{document}